\documentclass[11pt,a4paper]{article}
\usepackage[hyperref]{acl2020}
\usepackage{times}
\usepackage{latexsym}
\usepackage{linguex}
\usepackage{amsmath}
\usepackage{amssymb}
\usepackage{booktabs}
\usepackage{linguex}
\usepackage{url}
\usepackage{array}
\newcolumntype{x}[1]{>{\centering\arraybackslash\hspace{0pt}}p{#1}}
\newcolumntype{C}[1]{>{\centering\let\newline\\\arraybackslash\hspace{0pt}}m{#1}}
\usepackage{tabularx}
\usepackage{graphicx}
\usepackage{dblfloatfix}
\usepackage{enumitem}
\usepackage{ textcomp }
\usepackage{todonotes}
\usepackage{subcaption}
\usepackage{chngcntr}

\usepackage{microtype}
\usepackage{booktabs}

\aclfinalcopy 
\def\aclpaperid{718} 

\newlength{\vs}
\title{Syntactic Data Augmentation Increases\\ Robustness to Inference Heuristics}

\author{Junghyun Min\textsuperscript{1}\quad R. Thomas McCoy\textsuperscript{1}\quad Dipanjan Das\textsuperscript{2}\quad Emily Pitler\textsuperscript{2}\quad Tal Linzen\textsuperscript{1}\\
  \textsuperscript{1}Department of Cognitive Science, Johns Hopkins University, Baltimore, MD\\
  \textsuperscript{2}Google Research, New York, NY\\
  \texttt{\{jmin10, tom.mccoy, tal.linzen\}@jhu.edu}\\
  \texttt{\{dipanjand, epitler\}@google.com}}

\date{}

\begin{document}
\maketitle
\begin{abstract}
Pretrained neural models such as BERT, when fine-tuned to perform natural language inference (NLI), often show high accuracy on standard datasets, but display a surprising lack of sensitivity to word order on controlled challenge sets. We hypothesize that this issue is not primarily caused by the pretrained model's limitations, but rather by the paucity of crowdsourced NLI examples that might convey the importance of syntactic structure at the fine-tuning stage. We explore several methods to augment standard training sets with syntactically informative examples, generated by applying syntactic transformations to sentences from the MNLI corpus. The best-performing augmentation method, subject/object inversion, improved BERT's accuracy on controlled examples that diagnose sensitivity to word order from $0.28$ to $0.73$, without affecting performance on the MNLI test set. This improvement generalized beyond the particular construction used for data augmentation, suggesting that augmentation causes BERT to recruit abstract syntactic representations.
\end{abstract}

\section{Introduction}

In the supervised learning paradigm common in NLP, a large collection of labeled examples of a particular classification task is randomly split into a training set and a test set. The system is trained on this training set, and is then evaluated on the test set. Neural networks---in particular systems pretrained on a word prediction objective, such as ELMo \cite{peters2018elmo} or BERT \cite{devlin2019bert}---excel in this paradigm: with large enough pretraining corpora, these models match or even exceed the accuracy of untrained human annotators on many test sets \cite{raffel2019exploring}. 

\newlength{\myindent}
\setlength{\myindent}{0.5cm}

At the same time, there is mounting evidence that high accuracy on a \textit{test set} drawn from the same distribution as the training set does not indicate that the model has mastered the \textit{task}.
This discrepancy can manifest as a sharp drop in accuracy when the model is applied to a different dataset that illustrates the same task \cite{talmor2019multiqa,yogatama2019learning}, or as excessive sensitivity to linguistically irrelevant perturbations of the input \cite{jia2017adversarial,wallace2019universal}. 

One such discrepancy, where strong performance on a standard test set did not correspond to mastery of the task as a human would define it, was documented by \newcite{mccoy2019right} for the Natural Language Inference (NLI) task. In this task, the system is given two sentences, and is expected to determine whether one (the premise) entails the other (the hypothesis). Most if not all humans would agree that NLI requires sensitivity to syntactic structure; for example, the following sentences do not entail each other, even though they contain the same words:

\ex.The lawyer saw the actor.\label{lawyersubject}

\ex.The actor saw the lawyer.\label{lawyerobject}

\citeauthor{mccoy2019right} constructed the HANS challenge set, which includes examples of a range of such constructions, and used it to show that, when BERT is fine-tuned on the MNLI corpus \cite{williams2018multinli}, the fine-tuned model achieves high accuracy on the test set drawn from that corpus, yet displays little sensitivity to syntax; the model wrongly concluded, for example, that~\ref{lawyersubject} entails \ref{lawyerobject}.

We consider two explanations as to why BERT fine-tuned on MNLI fails on HANS.
Under the \textbf{Representational Inadequacy Hypothesis},
BERT fails on HANS because its pretrained representations are missing some necessary syntactic information.
Under the \textbf{Missed Connection Hypothesis}, BERT extracts the relevant syntactic information from the input (cf. \citealt{goldberg2019assessing,tenney2019context}), but it fails to \textit{use} this information with HANS because there are few MNLI training examples that indicate how syntax should support NLI \cite{mccoy2019right}. It is possible for both hypotheses to be correct: there may be some aspects of syntax that BERT has not learned at all, and other aspects that have been learned, but are not applied to perform inference.

The Missed Connection Hypothesis predicts that augmenting the training set with a small number of examples from one syntactic construction would teach BERT that the task requires it to use its syntactic representations. This would not only cause improvements on the construction used for augmentation, but would also lead to generalization to other constructions. In contrast, the Representational Inadequacy Hypothesis predicts that to perform better on HANS, BERT must be taught how each syntactic construction affects NLI from scratch. This predicts that larger augmentation sets will be required for adequate performance and that there will be little generalization across constructions.

This paper aims to test these hypotheses. We constructed augmentation sets by applying syntactic transformations to a small number of examples from MNLI. Accuracy on syntactically challenging cases improved dramatically as a result of augmenting MNLI with only about 400 examples in which the subject and the object were swapped (about $0.1\%$ of the size of the MNLI training set). Crucially, even though only a single transformation was used in augmentation, accuracy increased on a range of constructions. For example, BERT's accuracy on examples involving relative clauses (e.g, \textit{The actors called the banker who the tourists saw} $\nrightarrow$ \textit{The banker called the tourists}) was $0.33$ without augmentation, and $0.83$ with it. This suggests that our method does not overfit to one construction, but taps into BERT's existing syntactic representations, providing support for the Missed Connection Hypothesis. At the same time, we also observe limits to generalization, supporting the Representational Inadequacy Hypothesis in those cases.

\section{Background}

HANS is a template-generated challenge set designed to test whether NLI models have adopted three syntactic heuristics. First, the \textbf{lexical overlap heuristic} is the assumption that any time all of the words in the hypothesis are also in the premise, the label should be \textit{entailment}. In the MNLI training set, this heuristic often makes correct predictions, and almost never makes incorrect predictions. This may be due to the process by which MNLI was generated: crowdworkers were given a premise and were asked to generate a sentence that contradicts or entails the premise. To minimize effort, workers may have overused lexical overlap as a shortcut to generating entailed hypotheses. Of course, the lexical overlap heuristic is not a generally valid inference strategy, and it fails on many HANS examples; e.g., as discussed above, \textit{the lawyer saw the actor} does not entail \textit{the actor saw the lawyer}. 

HANS also includes cases that are diagnostic of the \textbf{subsequence heuristic} (assume that a premise entails any hypothesis which is a contiguous subsequence of it) and the \textbf{constituent heuristic} (assume that a premise entails all of its constituents). While we focus on counteracting the lexical overlap heuristic, we will also test for generalization to the other heuristics, which can be seen as particularly challenging cases of lexical overlap. Examples of all constructions used to diagnose the three heuristics are given in Tables~\ref{tab:subjobjgen_detailed_overlap},~\ref{tab:subjobjgen_detailed_subsequence} and~\ref{tab:subjobjgen_detailed_constituent}.

Data augmentation is often employed to increase robustness in vision \cite{perez2017effectiveness} and language \cite{belinkov2018synthetic,wei2019eda}, including in NLI \cite{minervini2018adversarially, yanaka2019help}. In many cases, augmentation with one kind of example improves accuracy on that particular case, but does not generalize to other cases, suggesting that models overfit to the augmentation set \cite{jia2017adversarial,ribeiro2018semantically,iyyer2018adversarial,liu2019inoculation}. In particular, \newcite{mccoy2019right} found that augmentation with HANS examples generalized to a different word overlap challenge set \cite{dasgupta2018evaluating}, but only for examples similar in length to HANS examples. We mitigate such overfitting to superficial properties by generating a diverse set of corpus-based examples, which differ from the challenge set both lexically and syntactically. Finally, \newcite{kim2018teaching} used a similar augmentation approach to ours but did not study generalization to types of examples not in the augmentation set.

\section{Generating Augmentation Data}
\label{sec:generating}

We generate augmentation examples from MNLI using two syntactic transformations: \textsc{inversion}, which swaps the subject and object of the source sentence, and \textsc{passivization}. For each  of these 
transformations, we had two families of augmentation sets. The \textsc{original premise} strategy keeps the original MNLI premise and transforms the hypothesis; and \textsc{transformed hypothesis} uses the original MNLI hypothesis as the new premise, and the transformed hypothesis as the new hypothesis (see Table~\ref{tab:sample_strategies} for examples, and \S\ref{sec:strategies_appendix} for details). We experimented with three augmentation set sizes: small ($101$ examples), medium ($405$) and large ($1215$). All augmentation sets were much smaller than the MNLI training set ($297k$).\footnote{The augmentation sets and the code used to generate them are available at \url{https://github.com/aatlantise/syntactic-augmentation-nli}.}

We did not attempt to ensure the naturalness of the generated examples; e.g., in the \textsc{inversion} transformation, \textit{The carriage made a lot of noise} was transformed into \textit{A lot of noise made the carriage}. In addition, the labels of the augmentation dataset were somewhat noisy; e.g., we assumed that \textsc{inversion} changed the correct label from \textit{entailment} to \textit{neutral}, but this is not necessarily the case (if \textit{The buyer met the seller}, it is likely that \textit{The seller met the buyer}).  As we show below, this noise did not hurt accuracy on MNLI.

Finally, we included a random shuffling condition, in which an MNLI premise and its hypothesis were both randomly shuffled, with a random label. We used this condition to test whether a  syntactically uninformed method could teach the model that, when word order is ignored, no reliable inferences can be made.

\setlength{\belowcaptionskip}{-2pt}
\begin{table}
    \centering
    \resizebox{0.95\columnwidth}{!}{
    \begin{tabular}{lll}
    \toprule 
        Original MNLI example:\\
        
        \hspace{\myindent} \textcolor{blue}{There are 16 El Grecos in this small collection.} \Large $\rightarrow$\\ 
        \hspace{\myindent} \textcolor{red}{This small collection contains 16 El Grecos.}\vspace{0.3cm}\\
        
        Inversion (original premise):\\
        
        \hspace{\myindent} \textcolor{blue}{There are 16 El Grecos in this small collection.} \Large $\nrightarrow$\\
        \hspace{\myindent}  16 El Grecos contain this small collection.\vspace{0.3cm}\\
        
        \multicolumn{3}{l}{Inversion (transformed hypothesis):}\\
        
        \hspace{\myindent}  \textcolor{red}{This small collection contains 16 El Grecos.} \Large $\nrightarrow$\\
        \hspace{\myindent}  16 El Grecos contain this small collection.\vspace{0.3cm}\\
        
        \multicolumn{3}{l}{Passivization (transformed hypothesis; \textit{non-entailment}):}\\
        
        \hspace{\myindent}  \textcolor{red}{This small collection contains 16 El Grecos.}
        \Large $\nrightarrow$ \\
        
        \hspace{\myindent}  This small collection is contained by 16 El Grecos.\vspace{0.3cm}\\
        \multicolumn{3}{l}{Random shuffling with a random label:}\\
        \hspace{\myindent}  16 collection small El contains Grecos This. 
        \Large $\nrightarrow$/$\rightarrow$\\
        \hspace{\myindent}  collection This Grecos El small 16 contains.\\
        \bottomrule
    \end{tabular}
}
\caption{A sample of syntactic augmentation strategies, with gold labels ($\rightarrow$: \textit{entailment}; $\nrightarrow$: \textit{non-entailment}). For the full list, see Table~\protect\ref{tab:full_strategies} in the Appendix.}\label{tab:sample_strategies}
\end{table}

\begin{figure*}
\centering
    \includegraphics[width=5.5in]{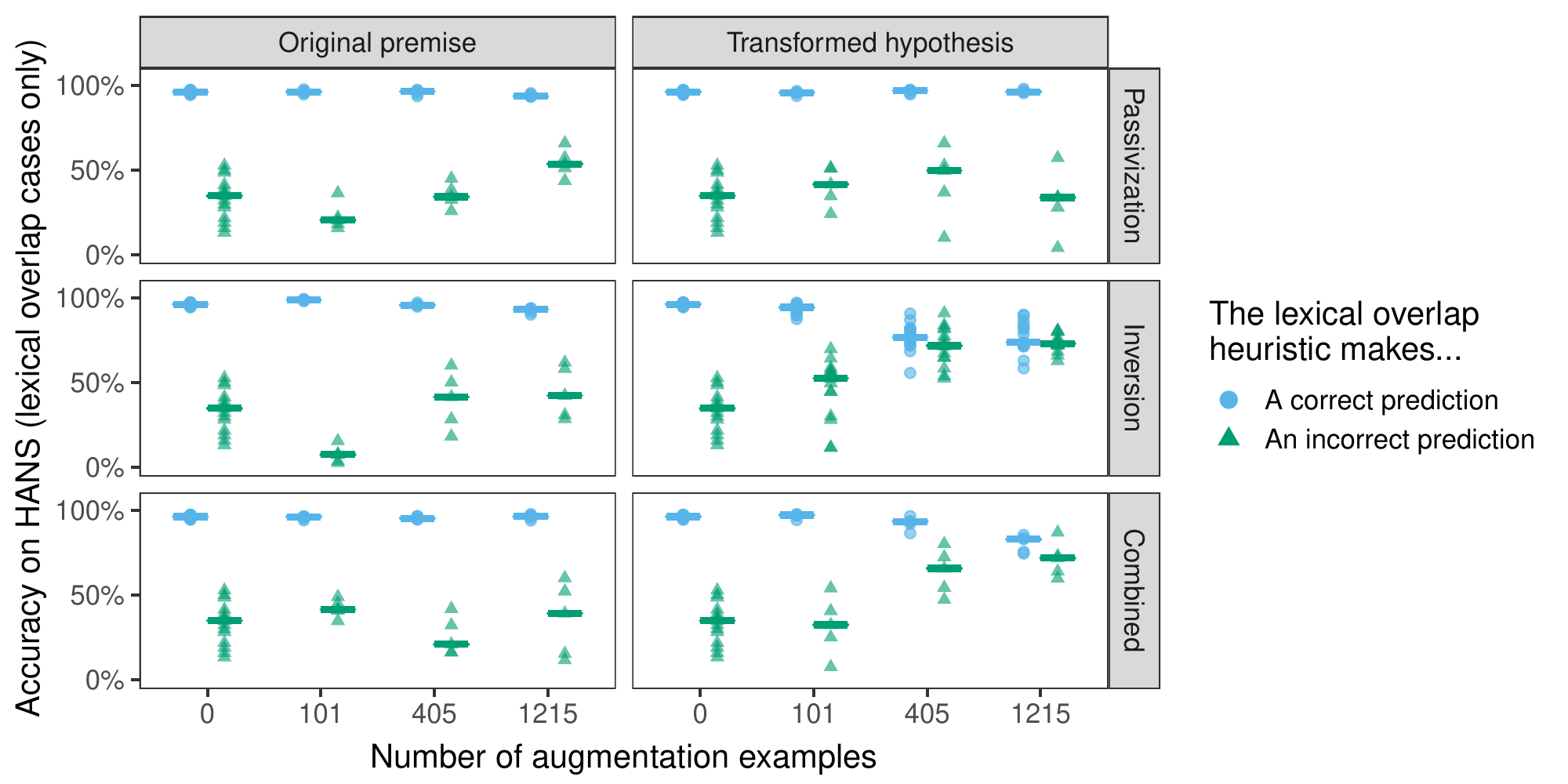}
    \caption{Comparison of syntactic augmentation strategies. Dots represent accuracy on the HANS examples that diagnose the lexical overlap heuristic, as produced by each of the runs of BERT fine-tuned on MNLI combined with each augmentation data set. Horizontal bars indicate median accuracy across runs. Chance accuracy is $0.5$.}\label{fig:strategy_accuracy_lexical}
\end{figure*}

\section{Experimental setup}

We added each augmentation set separately to the MNLI training set, and fine-tuned BERT on each resulting training set. Further fine-tuning details are in Appendix \ref{sec:fine_tuning_appendix}.
We repeated this process for five random seeds for each combination of augmentation strategy and augmentation set size, except for the most successful strategy (\textsc{inversion} + \textsc{transformed hypothesis}), for which we had 15 runs for each augmentation size. Following \newcite{mccoy2019right}, when evaluating on HANS, we merged the neutral and contradiction labels produced by the model into a single \textit{non-entailment} label.

For both \textsc{original premise} and \textsc{transformed hypothesis}, we experimented with using each of the transformations separately, and with a combined dataset including both inversion and passivization. We also ran separate experiments with only the passivization examples with an \textit{entailment} label, and with only the passivization examples with a \textit{non-entailment} label. As a baseline, we used 100 runs of BERT fine-tuned on the unaugmented MNLI \cite{mccoy2019berts}.

We report the models' accuracy on HANS, as well as on the MNLI development set (MNLI test set labels are not publicly available). We did not tune any parameters on this development set. All of the comparisons we discuss below are significant at the $p < 0.01$ level (based on two-sided t-tests).

\section{Results}

Accuracy on MNLI was very similar across augmentation strategies and matched that of the unaugmented baseline ($0.84$), suggesting that syntactic augmentation with up to $1215$ examples does not harm overall performance on the dataset. By contrast, accuracy on HANS varied significantly, with most models performing worse than chance (which is $0.50$ on HANS) on \textit{non-entailment} examples, suggesting that they adopted the heuristics (Figure~\ref{fig:strategy_accuracy_lexical}). The most effective augmentation strategy, by a large margin, was inversion with a transformed hypothesis. Accuracy on the HANS word overlap cases for which the correct label is \textit{non-entailment}---e.g., \textit{the doctor saw the lawyer} $\nrightarrow$ \textit{the lawyer saw the doctor}---was $0.28$ without augmentation, and $0.73$ with the large version of this augmentation set. Simultaneously, this strategy \textbf{decreased} BERT's accuracy on the cases where the heuristic makes the correct prediction (\textit{The tourists by the actor called the authors} $\rightarrow$ \textit{The tourists called the authors}); in fact, the best model's accuracy was similar across cases where lexical overlap made correct and incorrect predictions, suggesting that this intervention prevented the model from adopting the heuristic.

The random shuffling method did not improve over the unaugmented baseline, suggesting that syntactically-informed transformations are essential (Table \ref{tab:strategies_results_full}).  Passivization yielded a much smaller benefit than inversion, perhaps due to the presence of overt markers such as the word \textit{by}, which may lead the model to attend to word order only when those are present. Intriguingly, even on the passive examples in HANS, inversion was more effective than passivization (large inversion augmentation: $0.13$; large passivization augmentation: $0.01$). Finally, inversion on its own was more effective than the combination of inversion and passivization.

\begin{figure*}
\centering
    \includegraphics[width=5.5in]{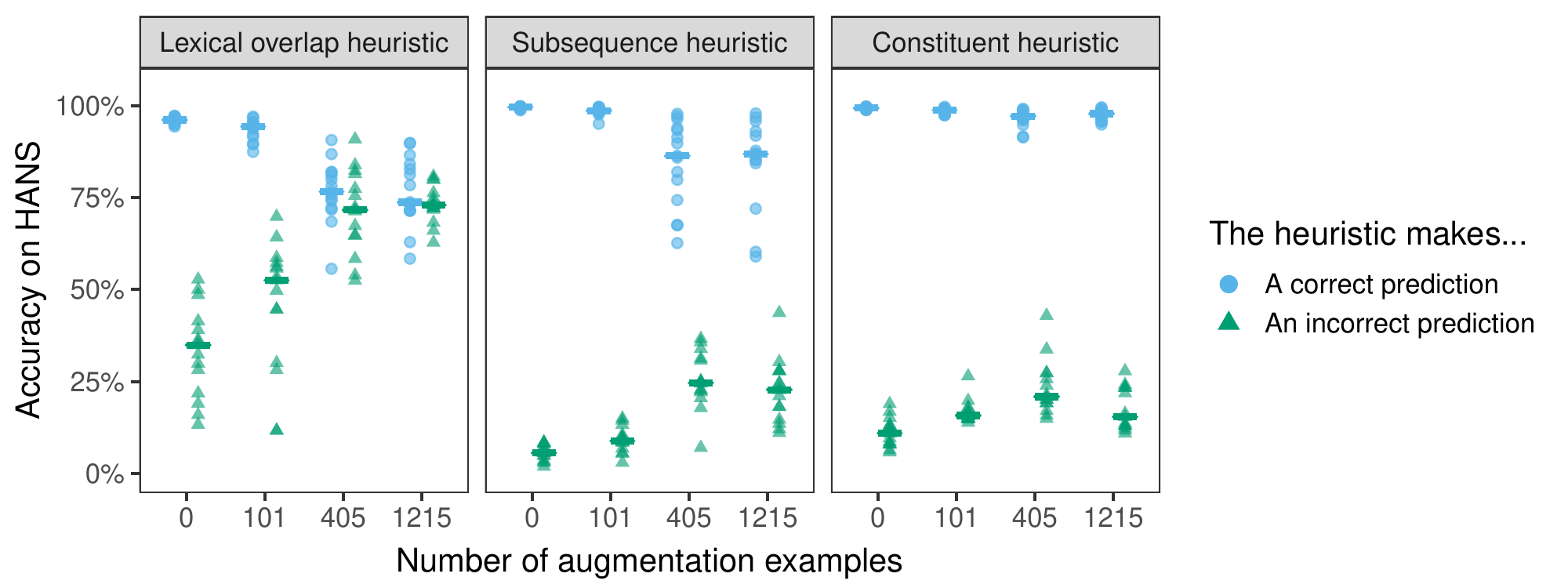}
    \caption{Augmentation using subject/object inversion with a transformed hypothesis. Dots represent the accuracy on HANS examples diagnostic of each of the heuristics, as produced by each of the 15 runs of BERT fine-tuned on MNLI combined with each augmentation data set. Horizontal bars indicate median accuracy across runs.}\label{fig:inversion_transformed}
\end{figure*}

We now analyze in more detail the most effective strategy, inversion with a transformed hypothesis. First, this strategy is similar on an abstract level to the HANS subject/object swap category, but the two differ in vocabulary and some syntactic properties; despite these differences, performance on this HANS category was perfect ($1.00$) with medium and large augmentation, indicating that BERT benefited from the high-level syntactic structure of the transformation. For the small augmentation set, accuracy on this category was $0.53$, suggesting that 101 examples are insufficient to teach BERT that subjects and objects cannot be freely swapped. Conversely, tripling the augmentation size from medium to large had a moderate and inconsistent effect across HANS subcases (see Appendix~\ref{sec:fine_grained} for case-by-case results); for clearer insight about the role of augmentation size, it may be necessary to sample this parameter more densely.

Although inversion was the only transformation in this augmentation set, performance also improved dramatically on constructions other than subject/object swap (Figure~\ref{fig:inversion_transformed}); for example, the models handled examples involving a prepositional phrase better, concluding, for instance, that \textit{The judge behind the manager saw the doctors} does not entail \textit{The doctors saw the manager} (unaugmented: $0.41$; large augmentation: $0.89$). There was a much more moderate, but still significant, improvement on the cases targeting the subsequence heuristic; this smaller degree of improvement suggests that contiguous subsequences are treated separately from lexical overlap more generally. One exception was accuracy on ``NP/S'' inferences, such as \textit{the managers heard the secretary resigned} $\nrightarrow$ \textit{The managers heard the secretary}, which improved dramatically from $0.02$ (unaugmented) to $0.50$ (large augmentation).  Further improvements for subsequence cases may therefore require  augmentation with examples involving subsequences. 

A range of techniques have been proposed over the past year for improving performance on HANS. These include syntax-aware models \cite{moradshahi2019hubert, pang2019improving}, auxiliary models designed to capture pre-defined shallow heuristics so that the main model can focus on robust strategies \cite{clark2019dont, he2019unlearn,mahabadi2019simple}, and methods to up-weight difficult training examples \citep{yaghoobzadeh2019robust}. While some of these approaches yield higher accuracy on HANS than ours, including better generalization to the constituent and subsequence cases (see Table~\ref{tab:compare_other_papers}), they are not directly comparable: our goal is to assess how the prevalence of syntactically challenging examples in the training set affects BERT's NLI performance, without modifying either the model or the training procedure.

\section{Discussion}
 
Our best-performing strategy involved augmenting the MNLI training set with a small number of instances generated by applying the subject/object inversion transformation to MNLI examples. This yielded considerable generalization: both to another domain (the HANS challenge set), and, more importantly, to additional constructions, such as relative clauses and prepositional phrases. This supports the Missed Connection Hypothesis: a small amount of augmentation with one construction induced abstract syntactic sensitivity, instead of just ``inoculating'' the model against failing on the challenge set by providing it with a sample of cases from the same distribution \cite{liu2019inoculation}.

At the same time, the inversion transformation did not completely counteract the heuristic; in particular, the models showed poor performance on passive sentences. For these constructions, then, BERT's pretraining may not yield strong syntactic representations that can be tapped into with a small nudge from augmentation; in other words, this may be a case where our Representational Inadequacy Hypothesis holds. This hypothesis predicts that pretrained BERT, as a word prediction model, struggles with passives, and may need to learn the properties of this construction specifically for the NLI task; this would likely require a much larger number of augmentation examples. 

The best-performing augmentation strategy involved generating premise/hypothesis pairs from a single source sentence---meaning that this strategy does not rely on an NLI corpus. The fact that we can generate augmentation examples from any corpus makes it possible to test if very large augmentation sets are effective (with the caveat, of course, that augmentation sentences from a different domain may hurt performance on MNLI itself).

Ultimately, it would be desirable to have a model with a strong inductive bias for using syntax across language understanding tasks, even when overlap heuristics leads to high accuracy on the training set; indeed, it is hard to imagine that a human would ignore syntax entirely when understanding a sentence.  An alternative would be to create training sets that adequately represent a diverse range of linguistic phenomena; crowdworkers' (rational) preferences for using the simplest generation strategies possible could be counteracted by approaches such as adversarial filtering \citep{nie2019adversarial}. In the interim, however, we conclude that data augmentation is a simple and effective strategy to mitigate known inference heuristics in models such as BERT.

\section*{Acknowledgments}

This research was supported by a gift from Google, NSF Graduate Research Fellowship No. 1746891, and NSF Grant No. BCS-1920924. Our experiments were conducted using the Maryland Advanced Research Computing Center (MARCC).

\bibliography{syntactic_augmentation}
\bibliographystyle{acl_natbib}

\clearpage
\appendix
\counterwithin{figure}{section}
\counterwithin{table}{section}

\section{Appendix}

\subsection{Fine-tuning details}\label{sec:fine_tuning_appendix}

We used \texttt{bert-base-uncased} for all experiments. As is standard, we fine-tuned this pretrained model on MNLI by training a linear classifier to predict the label from the CLS token's final layer embedding, while continuing to update BERT's parameters \cite{devlin2019bert}. The order of training examples was reshuffled for each model. All models were trained for three epochs.

\subsection{Generating augmentation examples}
\label{sec:strategies_appendix}

The following list describes the augmentation strategies we used. Table~\ref{tab:full_strategies} illustrates all of these strategies as applied to a particular source sentence. Note that inversion generally changes the meaning of the sentence (\textit{the detective followed the suspect} refers to a different event from \textit{the suspect followed the detective}), but passivization on its own does not (\textit{the detective followed the suspect} refers to the same event as \textit{the suspect was followed by the detective}).

\begin{itemize}
\item Inversion (original premise): For a source example $(p, h, \rightarrow)$, generate $(p, \textsc{inv}(h), \nrightarrow)$, where \textsc{inv} returns the source sentence with the subject and object switched. Ignore source examples whose label is $\nrightarrow$.
\item Inversion (transformed hypothesis): For a source $(p, h)$ (with any label), discard the premise $p$ and generate $(h, \textsc{inv}(h), \nrightarrow)$.
\item Passivization (original premise): For a source $(p, h)$ (with any label), generate $(p, \textsc{pass}(h))$, with the same label, where \textsc{pass} returns the passive version of the source sentence (without changing its meaning).
\item Passivization (transformed hypothesis): For a source $(p, h)$, discard the premise $p$, and generate two examples, one with an \textit{entailment} label---$(h, \textsc{pass}(h), \rightarrow)$---and one with a \textit{non-entailment} label---$(h, \textsc{pass}(\textsc{inv}(h)), \nrightarrow)$.
\end{itemize}

We identified transitive sentences in MNLI that could serve as source sentences using the constituency parses provided with MNLI, excluding the noisier \textsc{telephone} genre. We did so by searching for matrix S nodes with exactly one NP daughter of the VP, where the subject and the object were both full noun phrases (i.e., neither were a personal pronoun such as \textit{me}), and where the verb lemma was not \textit{be} or \textit{have}. We kept the original tense of the verb, and modified its agreement features if necessary (e.g., \textit{the movie stars Matt Dillon and Gary Sinise} was transformed into \textit{Matt Dillon and Gary Sinise star the movie}).

The size of the largest augmentation set was 1215 for all strategies. This size was determined based on the largest augmentation dataset we could generate from MNLI for the inversion with original premise strategy using the procedure mentioned above. For fair comparison, we kept the same size even for strategies where we could have generated a larger dataset. We also created a Medium dataset by randomly sampling 405 of the cases identifying using the procedure above, as well as a small dataset with 101 examples. We performed this process only once for each strategy: as such, runs varied only in the classifier's weight initialization and the order of examples but not in the augmentation examples included in training.

To create the Combined augmentation dataset, we concatenated the inversion and passivization datasets, then randomly discarded half of the examples (to match the size of the combined dataset with the others). As with the other datasets, we only did this once: the Combined augmentation set was the same across runs. One consequence of this procedure is that the number of passivization and inversion examples was not exactly identical.

\begin{table*}
    \centering
    \begin{tabular}{lll}
    \toprule 
        \textbf{Original}\vspace{\vs}\\
        \hspace{\myindent}\textcolor{blue}{There are 16 El Grecos in this small collection.} \Large $\rightarrow$\\ 
        \hspace{\myindent}\textcolor{red}{This small collection contains 16 El Grecos.}\\
        
        \midrule
        \textbf{Inversion}\vspace{\vs}\\
        
        Original premise:\\
        \hspace{\myindent}\textcolor{blue}{There are 16 El Grecos in this small collection.} \Large $\nrightarrow$\\
        \hspace{\myindent}16 El Grecos contain this small collection.\vspace{\vs}\\
        
        Transformed hypothesis:\\
        \hspace{\myindent}\textcolor{red}{This small collection contains 16 El Grecos.} \Large $\nrightarrow$\\
        \hspace{\myindent}16 El Grecos contain this small collection.\\
        
        \midrule
        \textbf{Passivization}\vspace{\vs}\\
        
        Original premise:\\
        \hspace{\myindent}\textcolor{blue}{There are 16 El Grecos in this small collection.} \Large $\rightarrow$ \\\vspace{\vs}
        \hspace{\myindent}16 El Grecos are contained by this small collection.\vspace{0.5\vs}\\
        
        Transformed hypothesis (\textit{entailment} label):\\
        
        \hspace{\myindent} \textcolor{red}{This small collection contains 16 El Grecos.}
        \Large $\rightarrow$ \\
        
        \hspace{\myindent} 16 El Grecos are contained by the small collection.\vspace{0.5\vs}\\

        Transformed hypothesis (\textit{non-entailment} label):\\
        
        \hspace{\myindent} \textcolor{red}{This small collection contains 16 El Grecos.}
        \Large $\nrightarrow$ \\
        
        \hspace{\myindent} This small collection is contained by 16 El Grecos.\\
        \midrule
        \textbf{Random shuffling} (with random label)\vspace{\vs}\\
        \hspace{\myindent}are collection. small El  this in 16 There Grecos 
        \Large $\nrightarrow$/$\rightarrow$\\
        \hspace{\myindent}collection This Grecos El small 16 contains.\\
        \bottomrule
    \end{tabular} 
    \caption{Syntactic augmentation strategies (full table).}
    \label{tab:full_strategies}
\end{table*}

\subsection{Detailed Results}
\label{sec:fine_grained}

The following tables provide the detailed results of our experiments. Table~\ref{tab:strategies_results_full} shows each strategy's mean accuracy on MNLI, as well on the HANS cases that diagnose each of the three heuristics (the Lexical Overlap Heuristic, the Subsequence Heuristic, and the Constituent Heuristic), for which the correct label is \textit{non-entailment} ($\nrightarrow$). Table~\ref{tab:subj_obj_gen} zooms in on the best-performing augmentation strategy---subject/object inversion with a transformed hypothesis---on BERT's accuracy on HANS, both when the correct label is \textit{entailment} ($\rightarrow$) and when the label is \textit{non-entailment} ($\nrightarrow$). Finally, the last three tables detail the effect of augmentation by inversion with a transformed hypothesis on each of the 30 HANS subcases, broken down by the heuristic that they were designed to diagnose: the Lexical Overlap Heuristic (Table~\ref{tab:subjobjgen_detailed_overlap}), the Subsequence Heuristic (Table~\ref{tab:subjobjgen_detailed_subsequence}), and the Constituent Heuristic (Table~\ref{tab:subjobjgen_detailed_constituent}). 

\begin{table*}[h]
\centering
\begin{tabular}{lccccccccccccc} \toprule
      & \multicolumn{3}{c}{MNLI} & \multicolumn{3}{c}{Overlap} & \multicolumn{3}{c}{Subsequence} & \multicolumn{3}{c}{Constituent} \\ 
\cmidrule(lr){2-4} \cmidrule(lr){5-7} \cmidrule(lr){8-10} \cmidrule(lr){11-13}
                    & {\it S} & {\it M} & {\it L} & {\it S}     & {\it M}      & {\it L}     & {\it S}     & {\it M}     & {\it L}    & {\it S} & {\it M} & {\it L}\\ \midrule
     \multicolumn{5}{l}{\textbf{Original premise}}\vspace{0.5\vs}\\   Inversion    & .84  & .84 & .84  & .07   & .40    & .44   & .01   & .06   & .12 & .06 & .09 & .12  \\
    Passivization  & .84  & .84 & .84 & .23   & .35    & .54   & .04   & .05   & .09  & .13 & .11 & .15 \\
             Combined       & .84  & .84 & .84  & .42   & .25    & .36   & .07   & .05   & .04 & .14 & .15 & .12  \\ \midrule
     \multicolumn{5}{l}{\textbf{Transformed hypothesis}}\vspace{0.5\vs}\\       Inversion & .84  & .84 & .84 & {\bf.46} & {\bf .71} & {\bf .73} & {\bf .09}   & {\bf .25}   & .23 & {\bf.17} & .23 & .18 \\ 
     Passivization  & .84  & .84 & .84 & .41   & .43    & .31   & .06   & .06   & .07 & .13 & .15 & .17  \\
                 Combined & .84  & .84 & .84 & .32   & .64    & .71   & .06   & .13   & {\bf .28} & .15 & {\bf.26} & {\bf.22}  \\
                 Pass. (only pos) & .84  & .84 & .84 & .30   & .20    & .29   & .04   & .04   & .05 &.10 &.13 &.11  \\ 
                 Pass. (only neg) & .84  & .84 & .85 & .36   & .45    & .39   & .06   & .06   & .06 & .15 & .13 & .13  \\ \midrule
     Random shuffling & .84  & .84 & .84 & .26 & .19 & .35  & .05 & .05 & .06 & .15 & .14 & .14\\
    \cmidrule(lr){2-4} \cmidrule(lr){5-7} \cmidrule(lr){8-10} \cmidrule(lr){11-13}
     Unaugmented & \multicolumn{3}{c}{.84} & \multicolumn{3}{c}{.28} & \multicolumn{3}{c}{.05} & \multicolumn{3}{c}{.13} \\
     \bottomrule
\end{tabular}
\caption{Accuracy of models trained using each augmentation strategy when evaluated on HANS examples diagnostic of each of the three heuristics---lexical overlap, subsequence and constituent---for which the correct label is \textit{non-entailment} ($\nrightarrow$). Augmentation set sizes are \textit{S} ($101$ examples), \textit{M} ($405$) and \textit{L} ($1215$). Chance performance is $0.5$.}\label{tab:strategies_results_full}
\end{table*}

\begin{table*}
\centering
\begin{tabular}{llcccc} \toprule
     Subset of HANS & Label           & Unaugmented     & Small     & Medium     & Large\\ \midrule
     MNLI & All             & 0.84   & 0.84   & 0.84   & 0.84\vspace{\vs}\\
     Subject/object swap & $\nrightarrow$   & 0.19   & 0.53   & 1.00   & 1.00\vspace{\vs}\\
     All other & $\rightarrow$    & 0.96   & 0.93   & 0.77   & 0.77 \\
     lexical overlap & $\nrightarrow$   & 0.30   & 0.44   & 0.64   & 0.66\vspace{\vs} \\
     Subsequence & $\rightarrow$ & 0.99  & 0.99   & 0.84   & 0.85 \\
     & $\nrightarrow$           & 0.05   & 0.09   & 0.25   & 0.23\vspace{\vs}\\
     Constituent & $\rightarrow$ & 0.99  & 0.98   & 0.97   & 0.97 \\
     & $\nrightarrow$           & 0.13   & 0.17   & 0.23   & 0.18\\
     \bottomrule
\end{tabular}
\caption{Effect on HANS accuracy of augmentation using subject/object inversion with a transformed hypothesis. Results are shown for BERT fined-tuned on the MNLI training set augmented with the three size of augmentation sets ($101$, $405$ and $1215$ examples), as well as for BERT fine-tuned on the unaugmented MNLI training set.}\label{tab:subj_obj_gen}
\end{table*}

\begin{table*}
\centering
\resizebox{\textwidth}{!}{
\begin{tabular}{lccccccc} \toprule
      & & \multicolumn{3}{c}{\textit{Entailment}} & \multicolumn{3}{c}{\textit{Non-entailment}} \\
      \cmidrule(lr){3-5} \cmidrule(lr){6-8}
     Architecture or training method & Overall & \textit{L} & \textit{S} & \textit{C} & \textit{L} & \textit{S} & \textit{C} \\ \midrule
     Baseline \cite{mccoy2019berts} & 0.57 & 0.96 & 0.99 & 0.99 & 0.28 & 0.05 & 0.13\vspace{\vs} \\
     Learned-Mixin + H \cite{clark2019dont} & 0.69 & 0.68 & 0.84 & 0.81 & 0.77 & 0.45 & 0.60 \\
     DRiFt-\textsc{HAND} \cite{he2019unlearn} & 0.66 & 0.77 & 0.71 & 0.76 & 0.71 & 0.41 & 0.61 \\
     Product of experts \cite{mahabadi2019simple} & 0.67 & 0.94 & 0.96 & 0.98 & 0.62 & 0.19 & 0.30 \\
     HUBERT + \cite{moradshahi2019hubert} & 0.63 & 0.96 & 1.00 & 0.99 & 0.70 & 0.04 & 0.11 \\
     MT-DNN + \texttt{LF} \cite{pang2019improving} & 0.61 & 0.99 & 0.99 & 0.94 & 0.07 & 0.07 & 0.13 \\
     BiLSTM forgettables \cite{yaghoobzadeh2019robust}& 0.74 & 0.77 & 0.91 & 0.93 & 0.82 & 0.41 & 0.61\vspace{\vs}\\
     \textbf{Ours:}\\
     Inversion (transformed hypothesis), small & 0.60 & 0.93 & 0.99 & 0.98 & 0.46 & 0.09 & 0.17 \\
     Inversion (transformed hypothesis), medium & 0.63 & 0.77 & 0.84 & 0.97 & 0.71 & 0.25 & 0.23 \\
     Inversion (transformed hypothesis), large & 0.62 & 0.77 & 0.85 & 0.97 & 0.73 & 0.23 & 0.18\\
     Combined (transformed hypothesis), medium & 0.65 & 0.92 & 0.96 & 0.98 & 0.64 & 0.13 & 0.26 \\
     \bottomrule
\end{tabular}
}
\caption{HANS accuracy from various architectures and training methods, broken down by the heuristic that the example is diagnostic of and by its gold label, as well as overall accuracy on HANS. All but MT-DNN + \texttt{LF} use BERT as base model. \textit{L}, \textit{S}, and \textit{C} stand for lexical overlap, subsequence, and constituent heuristics, respectively. Augmentation set sizes are \textit{n} = 101 for small, \textit{n} = 405 for medium, and \textit{n} = 1215 for large.}\label{tab:compare_other_papers}
\end{table*}

\clearpage

\begin{table*}
\centering
\begin{tabular}{p{5cm}ccccc} \toprule
     Subcase & Unaugmented & Small & Medium & Large\\ \midrule
     Subject-object swap & 0.19 & 0.53 & 1.00 & 1.00 \\
     \multicolumn{6}{l}{\textit{The senators mentioned the artist.}  $\nrightarrow$ \textit{The artist mentioned the senators.}} \vspace{\vs}\\
     Sentences with PPs & 0.41 & 0.61 & 0.81 & 0.89 \\
     \multicolumn{6}{l}{\textit{The judge behind the manager saw the doctors.}  $\nrightarrow$ \textit{The doctors saw the manager.}}\vspace{\vs} \\
     Sentences with relative clauses & 0.33 & 0.53 & 0.77 & 0.83 \\
     \multicolumn{6}{l}{\textit{The actors called the banker who the tourists saw.} $\nrightarrow$ \textit{The banker called the tourists.}}\vspace{\vs} \\
     Passives & 0.01 & 0.04 & 0.29 & 0.13 \\ 
     \multicolumn{6}{l}{\textit{The senators were helped by the managers.} $\nrightarrow$ \textit{The senators helped the managers.}}\vspace{\vs} \\
     Conjunctions & 0.45 & 0.59 & 0.69 & 0.81 \\
     \multicolumn{6}{l}{\textit{The doctors saw the presidents and the tourists.} $\nrightarrow$ \textit{The presidents saw the tourists.}}\\
     \midrule
     
     Untangling relative clauses & 0.98 & 0.94 & 0.74 & 0.76 \\
     \multicolumn{6}{l}{\textit{The athlete who the judges saw called the manager.} $\rightarrow$ \textit{The judges saw the athlete.}} \\
     \\
     Sentences with PPs & 1.00 & 0.98 & 0.85 & 0.86 \\
     \multicolumn{6}{l}{\textit{The tourists by the actor called the authors.}  $\rightarrow$ \textit{The tourists called the authors.}}\vspace{\vs} \\
     Sentences with relative clauses & 0.99 & 0.98 & 0.89 & 0.89 \\
     \multicolumn{6}{l}{\textit{The actors that danced encouraged the author.}  $\rightarrow$ \textit{The actors encouraged the author.}} \vspace{\vs} \\
     Conjunctions & 0.83 & 0.78 & 0.68 & 0.66 \\
     \multicolumn{6}{l}{\textit{The secretaries saw the scientists and the actors.} $\rightarrow$ \textit{The secretaries saw the actors.}}\vspace{\vs} \\
     Passives & 1.00 & 0.99 & 0.67 & 0.67  \\ 
     \multicolumn{6}{l}{\textit{The authors were supported by the tourists.} $\rightarrow$ \textit{The tourists supported the authors.}}\\

     \bottomrule
\end{tabular}
\caption{Subject/object inversion with a transformed hypothesis: results for the HANS subcases that are diagnostic of the lexical overlap heuristic, for four training regimens---unaugmented (trained only on MNLI), and with small ($n = 101$), medium ($n = 405$) and large ($n = 1215$) augmentation sets. Chance performance is $0.5$. Top: cases in which the gold label is \textit{non-entailment}. Bottom: cases in which the gold label is \textit{entailment}.} \label{tab:subjobjgen_detailed_overlap}
\end{table*}

\begin{table*}
\centering
\begin{tabular}{p{5cm}ccccc} \toprule
     Subcase & Unaugmented & Small & Medium & Large \\ \midrule
   
     NP/S & 0.02 & 0.03 & 0.47 & 0.50 \\
     \multicolumn{6}{l}{\textit{The managers heard the secretary resigned.} $\nrightarrow$ \textit{The managers heard the secretary.}}\vspace{\vs} \\ 
     PP on subject & 0.12 & 0.21 & 0.21 & 0.23 \\
     \multicolumn{6}{l}{\textit{The managers near the scientist shouted.} $\nrightarrow$ \textit{The scientist shouted.}}\vspace{\vs} \\
     Relative clause on subject & 0.07 & 0.13 & 0.14 & 0.13 \\
     \multicolumn{6}{l}{\textit{The secretary that admired the senator saw the actor.} $\nrightarrow$ \textit{The senator saw the actor.}}\vspace{\vs} \\
     MV/RR & 0.00 & 0.01 & 0.05 & 0.02 & \\
     \multicolumn{6}{l}{\textit{The senators paid in the office danced.} $\nrightarrow$ \textit{The senators paid in the office.}}\vspace{\vs} \\
     NP/Z & 0.06 & 0.09 & 0.41 & 0.25 \\ 
     \multicolumn{6}{l}{\textit{Before the actors presented the doctors arrived.} $\nrightarrow$ \textit{The actors presented the doctors.}} \\
     \midrule
     Conjunctions & 0.98 & 0.96 & 0.87 & 0.86 \\
     \multicolumn{6}{l}{\textit{The actor and the professor shouted.} $\rightarrow$ \textit{The professor shouted.}}\vspace{\vs} \\
     Adjectives & 1.00 & 1.00 & 0.92 & 0.91 \\
     \multicolumn{6}{l}{\textit{Happy professors mentioned the lawyer.} $\rightarrow$ \textit{Professors mentioned the lawyer.}}\vspace{\vs} \\
     Understood argument & 1.00 & 0.99 & 0.97 & 0.97 \\ 
     \multicolumn{6}{l}{\textit{The author read the book.} $\rightarrow$ \textit{The author read.}} \vspace{\vs} \\
     Relative clause on object & 0.99 & 0.98 & 0.70 & 0.71 \\ 
     \multicolumn{6}{l}{\textit{The artists avoided the actors that performed.} $\rightarrow$ \textit{The artists avoided the actors.}}\vspace{\vs} \\
     PP on object & 1.00 & 1.00 & 0.75 & 0.79 \\ 
     \multicolumn{6}{l}{\textit{The authors called the judges near the doctor.} $\rightarrow$ \textit{The authors called the judges.}}\\

     \bottomrule
\end{tabular}
\caption{Subject/object inversion with a transformed hypothesis: results for the HANS subcases diagnostic of the subsequence heuristic, for four training regimens---unaugmented (trained only on MNLI), and with small ($n = 101$), medium ($n = 405$) and large ($n = 1215$) augmentation sets. Top: cases in which the gold label is \textit{non-entailment}. Bottom: cases in which the gold label is \textit{entailment}.} \label{tab:subjobjgen_detailed_subsequence}
\end{table*}

\begin{table*}
\centering
\begin{tabular}{p{5cm}ccccc} \toprule
     Subcase & Unaugmented & Small & Medium & Large \\ \midrule

     Embedded under preposition & 0.41 & 0.43 & 0.57 & 0.49 \\
     \multicolumn{6}{l}{\textit{Unless the senators ran, the professors recommended the doctor.}  $\nrightarrow$ \textit{The senators ran.}}\vspace{\vs} \\
     Outside embedded clause & 0.00 & 0.01 & 0.02 & 0.01 \\
     \multicolumn{6}{l}{\textit{Unless the authors saw the students, the doctors resigned.} $\nrightarrow$ \textit{The doctors resigned.}}\vspace{\vs} \\
     Embedded under verb & 0.17 & 0.25 & 0.28 & 0.22 \\
     \multicolumn{6}{l}{\textit{The tourists said that the lawyer saw the banker.} $\nrightarrow$ \textit{The lawyer saw the banker.}}\vspace{\vs} \\
     Disjunction & 0.01 & 0.01 & 0.04 & 0.03 \\
     \multicolumn{6}{l}{\textit{The judges resigned, or the athletes saw the author.} $\nrightarrow$ \textit{The athletes saw the author.}}\vspace{\vs} \\
     Adverbs & 0.06 & 0.13 & 0.25 & 0.13 \\ 
     \multicolumn{6}{l}{\textit{Probably the artists saw the authors.}  $\nrightarrow$ \textit{The artists saw the authors.}}\\
     \midrule
     Embedded under preposition & 0.96 & 0.94 & 0.94 & 0.95 \\ 
     \multicolumn{6}{l}{\textit{Because the banker ran, the doctors saw the professors.} $\rightarrow$ \textit{The banker ran.}}\vspace{\vs} \\
     Outside embedded clause & 1.00 & 1.00 & 0.99 & 0.99 \\
     \multicolumn{6}{l}{\textit{Although the secretaries slept, the judges danced.} $\rightarrow$ \textit{The judges danced.}}\vspace{\vs} \\
     Embedded under verb & 0.99 & 0.99 & 0.98 & 0.97 \\
     \multicolumn{6}{l}{\textit{The president remembered that the actors performed.} $\rightarrow$ \textit{The actors performed.}}\vspace{\vs} \\
     Conjunction & 1.00 & 1.00 & 0.98 & 0.99 \\ 
     \multicolumn{6}{l}{\textit{The lawyer danced, and the judge supported the doctors.} $\rightarrow$ \textit{The lawyer danced.}}\vspace{\vs} \\
     Adverbs & 1.00 & 1.00 & 0.93 & 0.96 \\
     \multicolumn{6}{l}{\textit{Certainly the lawyers advised the manager.}  $\rightarrow$ \textit{The lawyers advised the manager.}}\\
     \bottomrule
\end{tabular}
\caption{Subject/object inversion with a transformed hypothesis: results for the HANS subcases diagnostic of the constituent heuristic, for four training regimens---unaugmented (trained only on MNLI), and with small ($n = 101$), medium ($n = 405$) and large ($n = 1215$) augmentation sets. Chance performance is $0.5$. Top: cases in which the gold label is \textit{non-entailment}. Bottom: cases in which the gold label is \textit{entailment}.}
\label{tab:subjobjgen_detailed_constituent}
\end{table*}

\end{document}